\documentclass[letterpaper, 10 pt, conference]{ieeeconf}  %
                                                          
\IEEEoverridecommandlockouts

\makeatletter\let\NAT@parse\undefined\makeatother %
\usepackage[numbers,sort&compress]{natbib}
\usepackage{amsmath,amsfonts}
\usepackage{algorithmic}
\usepackage{array}
\usepackage{textcomp}
\usepackage{stfloats}
\usepackage{url}
\usepackage{verbatim}
\usepackage{graphicx}
\usepackage{siunitx}
\usepackage{booktabs}
\usepackage{multirow}
\usepackage{pifont}

\usepackage[linesnumbered,vlined,ruled,commentsnumbered]{algorithm2e}
\SetKwRepeat{Do}{do}{while}

\SetVlineSkip{0.2em}

\SetInd{0.2em}{0.7em}

\DontPrintSemicolon{}

\SetKwIF{If}{ElseIf}{Else}{if}{}{else if}{else}{endif}

\SetAlgoSkip{}

\SetAlCapHSkip{0cm}

\SetAlgoCaptionLayout{small}
\SetAlFnt{\small}

\SetArgSty{textnormal}

\SetKwComment{Comment}{$\triangleright$\ }{}

\usepackage[tracking=true]{microtype}
\usepackage{amssymb}
\usepackage[colorlinks=true,citecolor=cyan,linkcolor=black,urlcolor=black,bookmarks=false,pdfusetitle]{hyperref}

\usepackage{pgfplots}
\pgfplotsset{compat=1.18}

\usepackage[capitalise]{cleveref}
\usepackage{xcolor}

\DeclareRobustCommand{\abbrevcrefs}{%
\crefname{algorithm}{Alg.}{Algs.}%
}
\abbrevcrefs

\usepackage{subcaption}
\begin{document}

\title{\LARGE \bf
\textsc{Wrap}: Fixtureless Wrench-aware Multi-Robot Assembly Planning}

\author{Valentin N. Hartmann, Huang Su, Yijiang Huang, Stelian Coros
\thanks{
All authors are with the Computational Robotics Lab, ETH Zurich, CH.}
}

\maketitle

\begin{abstract}
Assembly using robots often requires specially designed fixtures, or relies on top-down only assembly strategies.
Using multiple robots, we can avoid using fixtures and make robotic assembly more flexible.
Planning assembly sequences for multiple robots is challenging due to the high number of possible task assignments and orders.
In addition, we need to reason over forces that occur during the assembly process, e.g., to decide if multiple robots are required for support, or if external support such as a table should be used.

We present \textsc{Wrap}, a multi-robot assembly planner for multi-part assemblies, given the inter-part ordering-dependencies, the part meshes, and their initial state.
We formulate a linear program to reason about valid grasps for supporting the forces that occur during assembly.
The search leverages the assembly sequence, and greedily finds a feasible solution per assembly step by computing a heuristic via a cheap backwards search, and using the heuristic in the more expensive forward search.

We then solve the multi-robot, multi-goal motion planning problem, and for execution, we split the plan into contact-rich assembly skills, and free space motion.
We benchmark the planner on a variety of multi-part assemblies, and apply the planner to groups of robots differing in size and kinematics.
We validate the work both in a physics simulation, and in real.
Videos and code are available at \href{https://www.vhartmann.com/wrap}{\textls[-40]{\texttt{www.vhartmann.com/wrap}}}.
\end{abstract}

\section{Introduction}
Multi-part assemblies are all around us: be it things that we use in our day to day and assemble at home, such as furniture, or things that are assembled in factories, such as cars, or PCBs.
While robots are used for automated assembly in structured settings such as, building a car, assembly planning is tedious even in these scenarios, and relies on custom fixtures, manual task assignment, sequencing, and motion planning.

Recently, robots have also been applied to less structured assembly settings, but were usually limited by the required assumptions: To make the assembly tasks feasible for robots, possible restrictions are purely sequential assembly, pure top down assembly, or assuming access to a fixture to place the part and hold it in place.
Further, in most planners, the forces that happen during assembly are ignored, and realistic contact-rich assembly tasks (such as press fits, or screwing parts into another) might thus not be possible. %
By enabling flexible, fixtureless automated planning for high-mix low-repetition settings, robotic assembly becomes more economical \cite{marvel2018multirobot}.  

To illustrate what we aim to do, we begin with an example: Think of an instruction booklet for assembling a toy:
The instructions tell you the sequence of parts, and their goal location, but not how the part has to be grasped, or the path that your hands have to take in order to bring the part to the assembly pose.
The user needs to locate the part, decide how to reposition and possibly regrasp, and hold the subassemblies both in order to reach the assembly pose, and in order to support the forces that the assembly step requires.

\begin{figure}
    \centering
    \includegraphics[width=0.398\linewidth]{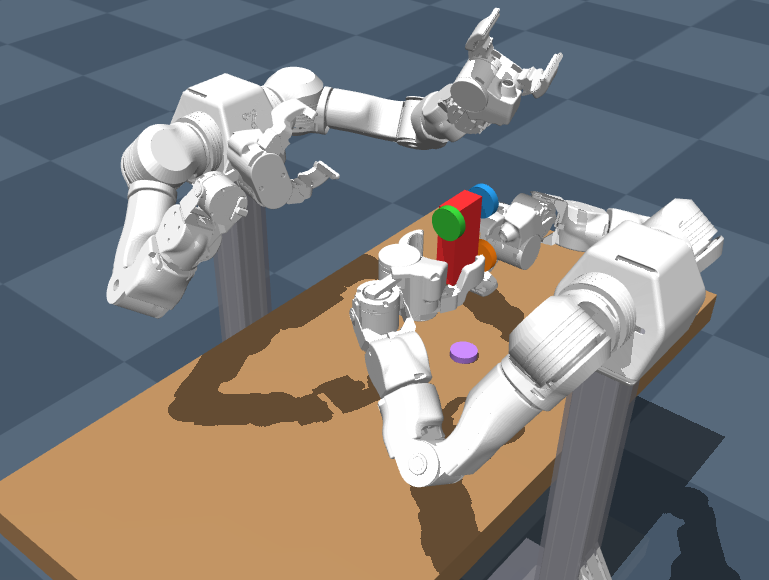}
    \includegraphics[width=0.45\linewidth]{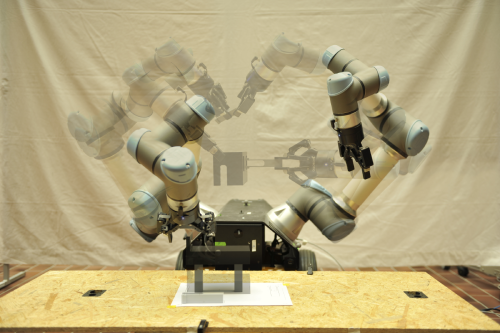}
    \caption{Multi-robot assembly: \textbf{(left)} Two OpenArm robots assembling a toy car. \textbf{(right)} Two UR5e arms collaborating to assemble a bench.}
    \label{fig:pull}
    \vspace{-3mm}
\end{figure}

To enable such behavior, we propose \textsc{Wrap}: a \textbf{W}rench-aware Multi-\textbf{R}obot \textbf{A}ssembly \textbf{P}lanning approach that does not require fixtures for assembly thanks to using multiple robots.
\textsc{Wrap} takes inter-part dependencies as input, and computes the required grasps and handovers in order to support the forces happening during the assembly process. 

We compare the plans from \textsc{Wrap} to baselines and present ablations.
We also compare the solutions from \textsc{Wrap} to plans without forces being taken into account in order to motivate why forces should be accounted for in the planning process.
We demonstrate the resulting motion plans both in a physics simulator and on a real robot system.
We open source the code, in addition to the 3D models that we used in this work.

\section{Related Work}

\subsection{Assembly Sequence Planning}
Assembly sequence planning determines an order and a corresponding motion to build a multi-part assembly.
Classical formulations are part-centric: parts are free-floating rigid bodies, and the questions are which subsets of parts can be separated~\cite{halperin2000general,jimenez2013survey,bahubalendruni2016review}.
Common approaches in this part-centric planning are sampling-based planning in the configuration space of the parts~\cite{sundaram2001disassembly,zhang2020cspace,bayraktar2026peel}, reinforcement learning for the combinatorial ordering problem~\cite{funk2022learn2assemble,ghasemipour2022blocks}, and physics simulation, which allows disassembling parts with tight clearances~\cite{tian2022assemble}.
If robots are considered at all, they usually enter as a feasibility filter: a sequence is rejected if a part is unreachable or if the partial assembly would collapse under gravity~\cite{tian2024asap,rodriguez2019iteratively}, or a contact is selected once the sequence has been inferred~\cite{zhu2024multilevel}.
Our work is complementary: we take a dependency graph as input, and reason over the robot action sequence: which robot grasps which part with which grasp, when parts are handed over or set down, and how is the mating wrench balanced are decided within one search, since none of these questions can be answered by reasoning about part motions alone.

\paragraph*{Forces and Stability in Assembly Planning}
Mattikalli et al.~\cite{mattikalli1995gravitational,mattikalli1996stable} formulate the gravitational stability of a partial assembly with friction as a linear program over contact wrenches, \cite{rakshit2015influence} extends the analysis along the motion path of a single gripper, and \cite{moriyama2019dualarm} adds such gravitational constraints to a dual-arm assembly planner.
In task and motion planning, forces have been incorporated as decision variables~\cite{20-toussaint-RAL}, and Holladay et al.~\cite{holladay2024robust} plan forceful manipulation by verifying that a single forceful kinematic chain can sustain the task wrench robustly.
To the best of our knowledge, no prior work considers forceful multi-robot assignment and sequencing.

\subsection{Multi-Robot Assembly Systems and Planning}
Robotic assembly has also been realized as complete systems in specific domains, from IKEA chairs~\cite{suarez2018can} and LEGO models~\cite{nagele2020legobot} to building-scale structures \cite{huang2021robotic,wang2023temporal,liang2017ras}.
Each of these systems is engineered around the geometry and joints of its own domain and does not transfer to assemblies of another kind.
In manufacturing, using a second robot as a ``dynamic fixture'' is an established strategy to avoid part-specific tooling~\cite{marvel2018multirobot}, but the holder and inserter roles are usually fixed. %
Dogar et al.~\cite{dogar2019multi} solve multi-robot grasp selection for a given assembly sequence as a constraint satisfaction problem, but optimistically assume that any required reorientation can be attained by handovers without explicitly planning them, and forces are entirely disregarded.
Long-horizon planners~\cite{22-hartmann-TRO,chen2022coop,hargus2026coordinated} compute robot assignments, sequences, and coordinated paths from a dependency structure, but all of these abstract the mating operation away.
Complementarily, AutoMate learns policies for contact-rich mating motions across diverse geometries~\cite{tang2024automate}, but does not address multi-part sequencing, robot assignment, or force-support planning.

Closest to our work, Fabrica~\cite{tian2025fabrica} integrates sequence, grasp, and motion planning for dual-arm assembly of general multi-part objects and adds supporting actions to counter-balance the insertion force of the other robot.
However, the partial assembly is held rigidly at a fixed pose once assembled; regrasping and reorientation is disallowed, which limits the reachability of the system and the types of parts it can assemble, and mating forces are not modeled beyond a grasp-stability heuristic.
\textsc{Wrap} plans handovers, regrasps, and set-downs of subassemblies without fixtures, and verifies for every step which agents can balance the mating wrench.

\section{Problem formulation}
An assembly is given as a set of parts $\mathcal{P}$ with initial poses $p_i^0$, and a set of valid final relative poses $P_i^f$ with respect to the part they are mated onto. 
We allow multiple possible assembly orientations since some parts are symmetric. 
We use $\mathcal{A} \subseteq 2^\mathcal{P}\setminus \{\varnothing\}$ for the set of all possible subassemblies.
Similar to \cite{dogar2019multi} we then describe the assembly sequence as the dependency graph $G = (\mathcal{A}, E)$ between assembly steps.
The vertices are the subassemblies (or parts) $A_i \in \mathcal{A}$, and edges $(A_i, A_j)\in E$ connect a preceding subassembly to the subassembly it is used to build.
One assembly step then takes (multiple) sub-assemblies as input (all parent vertices of $A_j$), and produces one new subassembly $A_j$, thus enabling first building (multiple) sub-assemblies that are later assembled.

Parts are mated (e.g., inserted or screwed) onto other parts with a controller that applies a maximum nominal mating wrench $w_i$.
The (new) connection between two assembled parts $p$ and $q$ is not perfectly rigid, but can only sustain a convex set of wrenches $\mathcal{T}_{p,q}$.
Assembly actions then need to fulfill both a collision constraint, where the robots and their end effectors avoid each other, and the combined wrench capabilities of the grasps need to fulfill a wrench constraint arising from the required assembly force.

For execution, we have a heterogeneous set of robots $\mathcal{R}$ with composite configuration space $\mathcal{Q}$, and end effectors with wrench capabilities $\mathcal{W}_r$.
Additionally, the table has friction, and we model the wrench that the table can balance using the wrench set $\mathcal{W}_\text{table}$.

The problem that we aim to solve is then finding the discrete action sequence $S$ and its parameterization, i.e., when, how, and which robot manipulates which object, and how assemblies need to be supported.
With this multi-robot action sequence, we then compute configurations that fulfill the constraints imposed at task boundaries by the action sequence.
Finally, we need to find a collision free trajectory $\pi(t):\mathbb{R}\rightarrow\mathcal{Q}$ that includes the execution of the controllers for the assembly actions.

\section{Multi-robot assembly planning}

\begin{figure*}
    \centering
    \includegraphics[width=0.9\linewidth]{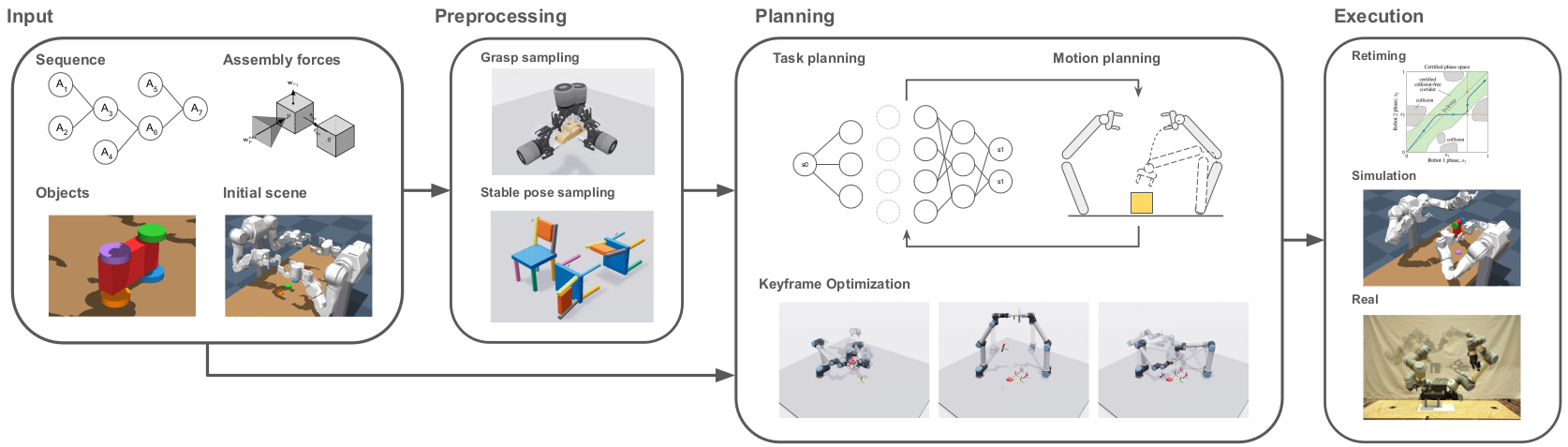}
    \caption{System overview: The input to our algorithm is part meshes, the initial state of the scene, and assembly dependencies, and mating wrenches. It then computes grasps, and stable rest orientations. After, it plans discrete robot-action sequences and their corresponding continuous configurations. The plan is then checked by a kinematic motion planner, and depending on feasibility of the path, replanning the plan is required. We then optimize the keyframes, and finally, we postprocess the path to make it dynamically feasible, and execute it. 
    }
    \label{fig:overview}
\end{figure*}

Our approach is described in \cref{alg:wrap} and an overview is given in \cref{fig:overview}.
Informally, we formulate the problem of finding a feasible action sequence as a search over a set of robot actions and their parameterizations.
In this work, we only consider prehensile manipulation. %

\textsc{Wrap} computes a set of valid grasps $\mathcal{G}_{A_i}$ using a grasp sampler, and stable placement poses $\mathcal{O}_{A_i}$ for all subassemblies $A_i$ appearing in the dependency graph $G$, then runs the search, and computes a motion plan for the found action plan using \cite{hartmann2025benchmark}.
Finally, the path is postprocessed to fulfill dynamics constraints, and the contact-rich segments are replaced by local controllers that deal with interaction forces.

\begin{algorithm}[t]
\caption{High level search of \textsc{Wrap}.}
\label{alg:wrap}
\KwIn{Assembly problem $\mathcal A$, flag \texttt{optimize}}
\KwOut{Task plan $\pi^\star$ and motion $\xi^\star$}
$(\mathcal O,\mathcal G)\gets\textsc{Preprocess}(\mathcal A)$\;
$\mathcal S\gets\textsc{InitializeTaskSearch}(\mathcal A,\mathcal O,\mathcal G)$\;
$\mathcal Q\gets\emptyset$,\ $I\gets\bot$\;
\While{budget remains and candidates remain}{
    $\pi\gets\textsc{NextTaskPlan}(\mathcal S,I)$\;
    \If{$\pi\neq\bot$}{
        $\mathcal Q\gets\mathcal Q\cup\{(\pi,\textsc{SelectWitnesses}(\pi))\}$\;
    }
    $(\pi,W)\gets\textsc{SelectFairly}(\mathcal Q)$\;
    $\xi\gets\textsc{MotionPlan}(\pi,W,\textsc{NextEffort}(\pi,W))$\;
    \eIf{$\xi\neq\bot$}{
        $I\gets\textsc{KeepBest}(I,(\pi,W,\xi))$\;
        \lIf{\textnormal{not \texttt{optimize}}}{\Return $(\pi,\xi)$}
    }{
        \textsc{Suspend}($\pi,W$)\;
    }
}
\lIf{$I=\bot$}{\Return \textsc{Unknown}}
\Return \textsc{Optimize}($I$)\;
\end{algorithm}

\subsection{Force constraints}
\begin{figure}
    \centering
    \includegraphics[width=0.9\linewidth]{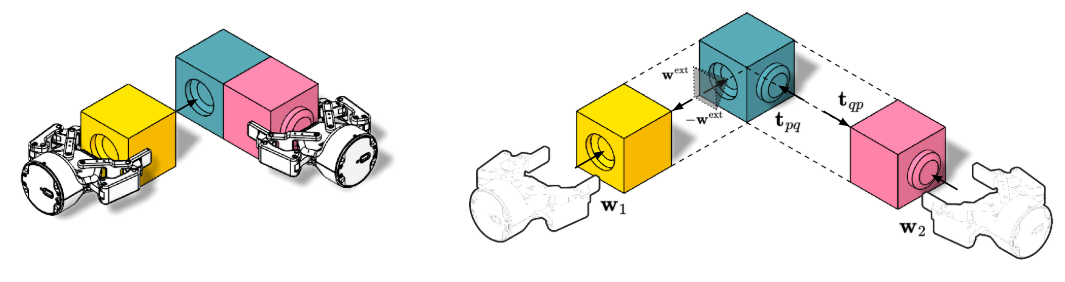}
    \caption{\textbf{(left)} Illustration of an assembly action and \textbf{(right)} the force constraints that we introduce.}
    \label{fig:force_eq}
\end{figure}

To evaluate if a certain assembly step is feasible, we solve a linear program (LP) to check if a static equilibrium (\cref{fig:force_eq}) exists for a set of grasp capabilities $\mathcal{W}_r$, and possible wrench capabilities $\mathcal{T}_{p,q}$ per part:
\begin{equation}
\label{eq:per_part_eq}
\sum_{i\in\mathcal{R}(p)\cup \{\text{table}\}}\mathbf{w}_i
+
\sum_{q\in\mathcal{N}(p)}\mathbf{t}_{qp}
+
\mathbf{w}^{\mathrm{ext}}_p
=
\mathbf{0},
\quad \forall p\in\mathcal{P},
\end{equation}
where $p$ is the part we are currently considering, $\mathcal{R}$ is the set of robots acting on part $p$, $\mathcal{N}$ is the set of parts that are connected to $p$, and $\mathbf{w}_p^\text{ext}$ is the external wrench acting on part $p$.
Gravitational forces are contained in $\mathbf{w}_p^\text{ext}$, and we model table support as a wrench with capability set $\mathcal{W}_\text{table}$, that is active if the part is on the table.
The LP is then
\begin{equation}
\label{eq:wrench-feasibility}
\begin{aligned}
\exists \quad & \mathbf{w}_\text{table}, \{\mathbf{w}_r\}_{r\in\mathcal{R}}, \{\mathbf{t}_{pq}\}_{(p,q)\in\mathcal{E}}
\\
\text{s.t.} \quad &
\cref{eq:per_part_eq},\\
&
\mathbf{w}_r \in \mathcal{W}_r, \ \ \ \ \forall r\in\mathcal{R}, \ \ \ \ \ \ 
\mathbf{w}_t \in \mathcal{W}_\text{table}, \\&
\mathbf{t}_{pq} \in \mathcal{T}_{p,q}, \ \ \ \forall (p,q)\in\mathcal{E},
\end{aligned}
\end{equation}
where $\mathcal{E}$ is the set of interfaces between parts.

\subsubsection*{Robust satisfaction}
The forces happening during an assembly step will not be perfectly what we expect at planning time, e.g., due to observation noise, minor grasp errors, or imperfect controllers.
We model the additional forces as the nominal force, plus bounded components along the transverse directions, and a positional offset to the application point of the force.
With gravity, this leads to a set of external wrenches $\mathcal{U}$ that needs to be satisfied.
We then require solving the LP for all the vertices of $\mathcal{U}$:
\begin{equation}
\forall\,
\mathbf{w}^{\mathrm{ext}}
\in
\text{vert}(\mathcal{U}),
\quad
\exists\,
\{\mathbf{w}_r\},
\{\mathbf{t}_{ij}\}
\quad
\text{s.t.}
\quad
\eqref{eq:wrench-feasibility}.
\end{equation}

With this robust satisfaction we see behavior such as placing a component on the table to resist the nominal press, while adding a brace to withstand possible lateral deviations that could otherwise topple it.

\subsection{Preprocessing}
Before we can proceed with task planning, we need to compute stable resting orientations $\mathcal{O}_{A_i}$ and stable grasps $\mathcal{G}_{A_i}$ for all subassemblies $A_i$ appearing in the assembly process, and for all involved end effector types.

\subsubsection*{Stable placement orientations}
To place objects back on the working surface after picking them up, we compute stable resting poses $\mathcal{O}_{A_i}\subset\text{SO}(3)$ for each subassembly by finding orientations that lead to the center of mass being inside of the supporting points of the convex hull of the subassembly.

\subsubsection*{Grasp sampling}
To manipulate the parts that we eventually want to assemble, we need to find a set of stable grasps $\mathcal{G}_{A_i}$ for all parts and eventual subassemblies.
We compute grasps via antipodal sampling and check that there is a certain minimum overlap between the finger pads and the object.
This could be replaced by a grasp generator, e.g.,  \cite{fang2023anygrasp}, which could also check in simulation if the grasp is stable under a given load.
For the vacuum gripper, we sample grasps that are normal to the surface, and have a minimum area of support.

\subsection{Task planning}
\begin{algorithm}[t]
\caption{Lazy milestone fill of \textsc{Wrap}.}
\label{alg:fill}
\KwIn{State $s$, milestone $m$}
\KwOut{Validated milestone plan $\rho$}
$h_m\gets\textsc{ReverseAbstractDijkstra}(m)$\;
\ForEach{nested action set $\Phi$}{
    \While{search or IK effort remains}{
        $\rho\gets\textsc{BestFirstFill}(s,m,h_m,\Phi)$\;
        \lIf{$\rho=\bot$}{\textbf{break}}
        $\textit{valid}\gets\texttt{true}$\;
        \ForEach{$e\in\textsc{UnvalidatedEdges}(\rho)$}{
            $\ell\gets\textsc{NextEffortLevel}(e)$\;
            \eIf{\textsc{ValidateIK}(e,$\ell$)}
                {\textsc{CacheWitness}(e)\;}
                {\textsc{SuspendEdge}(e,$\ell$)\;$\textit{valid}\gets\texttt{false}$\;\textbf{break}\;}
        }
        \lIf{\textit{valid}}{\Return $\rho$}
    }
    \textsc{ReopenAndWiden}($\Phi$)\;
}
\Return $\bot$\;
\end{algorithm}

We formulate a search problem over a set of actions, and their discretized parameterizations.
The actions we consider are pick/place/assemble/handover/add grasp/release, and their parameterizations are, e.g., the ID of a grasp transformation, or the ID of a placement pose, and which robot(s) are doing the task.
The state that we search over is then
\begin{equation}
    S_\text{fwd} = \Big(
        \overbrace{\mathstrut C}^{\text{\footnotesize assembly status}},
        \overbrace{\mathstrut\{(A_i,\sigma_i)\}_i}^{\text{\footnotesize Table objs.}},
        \overbrace{\mathstrut\{(r_j, p_j, \kappa_j)\}_j}^{\text{\footnotesize Grasped objs.}}
    \Big)
\end{equation}
where $C \subseteq \mathcal{A}$ denotes the set of subassemblies that have already been completed, $\sigma_i$ is the id of the placement pose of assembly $A_i$, and $\kappa$ the id of the grasp transformation with which robot $r$ grasps part $p$.
Each action has a pre- and a post-condition, which means that from a given initial symbolic state $s\in S_\text{fwd}$, we apply the actions to alter the state.
We can then run a search in order to find a valid action sequence that brings us to the desired final assembly.
However, a standard optimal search over the full state space $S_\text{fwd}$ does not scale both due to the branching, and the expensive validity checks.

We leverage the given dependency graph of parts to split the search in milestones, where we only compute robot action sequences for one assembly step at a time.
In this milestone search, we backtrack when we fail to find a solution, and try another part if possible, or report the problem as infeasible.
To manage the high branching factor due to many grasps, and many possible placements, we perform the search with a subset of the grasps and placement poses, which is widened if no solution could be found at the current resolution level. 
The search (\cref{alg:fill}) for a concrete robot action sequence is then split into a forward, and a backward component.

\subsubsection*{Backward Search}
In the backward pass, we compute a heuristic by disassembling the parts.
We only do cheap feasibility checks, such as collision checking of grasps by placing the involved end effectors at the grasp positions, and checking the wrench LP.
However, even a backward pass with only cheap checks runs in to the branching issues described before.
Thus, we run the search over a projected state %
\begin{equation}
    S_{\text{bwd}} = \Big(
        \overbrace{\mathstrut a}^{\text{\footnotesize status}},
        \overbrace{\mathstrut
            \{(p_j,K_j)\}_{j}
        }^{\text{\footnotesize assembly grasps}},
        \overbrace{\mathstrut \phi}^{\text{\footnotesize table rest}},
        \overbrace{\mathstrut
            \{(p_\ell,K_\ell)\}_{\ell}
        }^{\text{\footnotesize carried objs.}}
    \Big).
\end{equation}
Here, $a$ indicates if the current assembly step is finished, the tuple $(p_j, K_j)$ describes how part $p_j$ of the current target assembly is currently grasped, $\phi$ indicates if and how the assembly rests on the table, and finally, $(p_\ell, K_\ell)$ describes which loose components are carried, and how.
Crucially, the grasp labels $K$ are not the original grasp identifiers, and $\phi$ is not the original placement identifier.
Instead, they identify equivalence classes that preserve only some information of the original grasp or placement.

Empirically, we found that a good tradeoff between the size of the search space, and the informativeness of the projection is `can this grasp transmit the required assembly wrench', i.e., we project each grasp id $\kappa$ onto this binary class.

The heuristic is then computed by running a Dijkstra search over the state, where the actions are the `reverse' of the forward actions.
Concretely, we initialize all valid abstract goal states with distance zero and repeatedly apply inverse forward actions.
Whenever an action refers to a grasp class, it is considered feasible if \textit{any} grasp represented by that class satisfies the corresponding geometric and wrench constraints.

\subsubsection*{Forward Search}
In the forward search, we use this heuristic by projecting the forward state to its backwards equivalence class.
We run a greedy lazy shortest path search, and do inverse kinematics to try to find a valid configuration for the actions.

\subsubsection*{Plan improvement}
Once a valid task plan was found, we continue running the search in order to improve the path:
We run a large neighborhood search (LNS) \cite{Pisinger2019}, where in each fill window, we first run a greedy search, and then an exact A* search, using the previously found plan as upper bound on the cost in the forward and in the backward search.
We then widen the fill window (i.e., assemble $n$ parts in one search), and repeat the process until the window size is the full assembly horizon, or the timeout is reached.
This improves the robot action sequence, but we keep the initial assembly sequence that we chose, i.e., our task planner is not optimal for non-sequential assemblies.

\subsubsection*{Inverse kinematics (IK)}
The constraints for the IK problem we solve are given by the action we are considering: for a handover, all involved robots need to be at the selected grasping position, but the pose of the object is free; for a place action, the pose and the grasp is given, and only the configuration of the arm is free.
In addition to the constraints from the poses that need to be reached, we also enforce a minimum clearance between robots.

We use an optimization-based IK solver that we initialize with analytical solutions, and restart a limited amount of times with different seeds.
This allows a slight relaxation of the hard grasp constraint, which in turn allows using fewer grasps in the search.
However, an optimization-based solver can never certify infeasibility, so we do not mark a search node as infeasible based on the lack of an IK solution, but store it in a queue to revisit later at a different effort level \cite{toussaint2024effort}.
We use the number of restarts as `effort' here.

\subsubsection*{Joint Keyframe Optimization}
After a full task plan was found, we jointly optimize over sets of keyframes.
We do so by generating candidates for each keyframe, and then use dynamic programming to select a sequence minimizing IK-branch changes, and joint motion.
We additionally maximize per-keyframe clearance between robots in order to avoid configurations where robots are closely intertwined.
This step does not change the task sequence.

\subsection{Motion planning}
We formulate the motion planning problem as a multi-modal, multi-goal, multi-robot path planning problem, which we can solve using \cite{hartmann2025benchmark}.

A task plan might not have a feasible motion plan.
Hence, we include a feedback loop from motion planning to task planning. %
Similar to IK, our motion planner can not certify infeasibility, thus we suspend motion planning on timeout, store the task plan, and invest more planning effort later. %

\subsection{Retiming and Execution}
We postprocess the motion plan in order to satisfy kinodynamic constraints of the robots, to enable execution on real hardware.
We retime the paths: each robot stays constrained to its geometric path, but may advance along it independently.
The manipulation events, and possible collisions if deviating too far from the original timing impose synchronization constraints for this independent optimization.
We reserve expected skill durations, allowing unrelated robots to continue advancing on their paths up until a certain maximum deviation.
If a manipulation skill exceeds its reserved duration, the participating robots wait at the event while unrelated continue until they are synchronized again.
In contrast to APEX-MR~\cite{huang2025apexmr}, which represents asynchronous execution using a temporal plan graph, we directly optimize and certify the robots' continuous path phases.

\section{Experiments}
\subsection{Benchmark Suite and Setup}
\begin{figure}[t]
    \centering
    \includegraphics[width=0.9\linewidth]{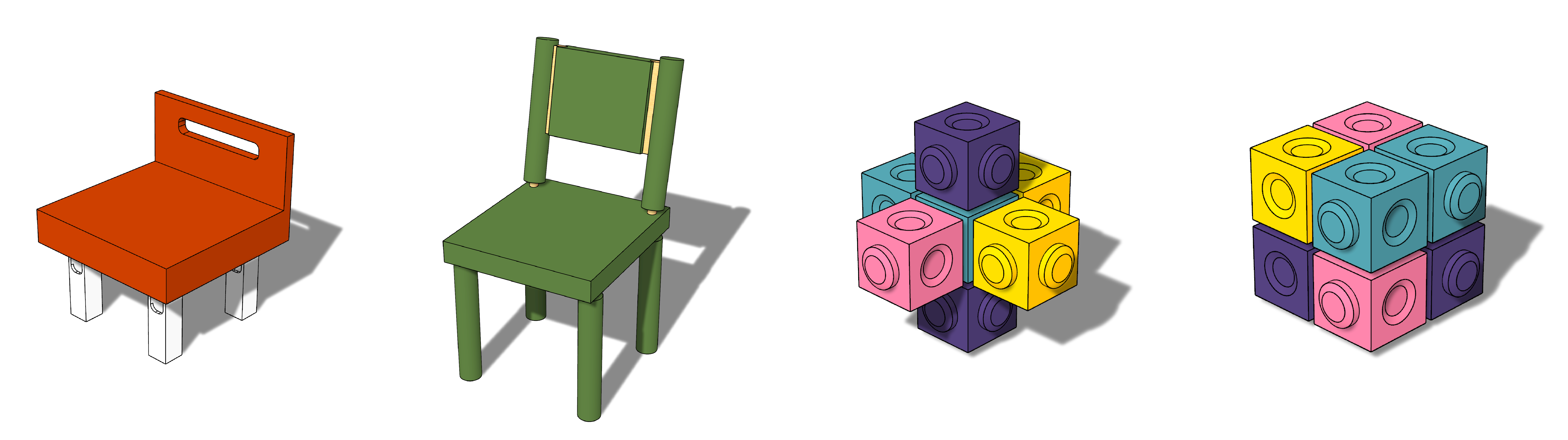}
    \caption{New assemblies from left to right: Stool, Chair, Cross, Cube.}
    \label{fig:assemblies}
\end{figure}

We introduce four new assemblies (\cref{fig:assemblies}): 
\begin{itemize}
    \item Stool (5 parts): A stool with a base and 4 legs, which we use for testing the assembly plan in real. The legs have a snap-fit mechanism.
    \item Chair (8 parts): a chair consisting of a base with 4 legs, and a backrest, which has to be assembled separately before both modules can be assembled.
    \item Cross (7 parts): A cross-like shape in 3D, which requires many reorientations to add all cubes onto the center cube, which we use for the scaling analysis.
    \item Cube (8 parts): A 2x2x2 cuboid consisting of smaller cubes, which we only use for the scaling analysis. 
\end{itemize}

We also compare against the objects that were introduced in \cite{tian2025fabrica} (Beam (5 parts), Plumber block (5), Car (5), Gamepad (6), Cooling Manifold (7), Duct (8), and Stool (9)), but remove the fixtures for the initial poses of the parts, and the fixtures that allow on-table assembly without any reorientation.
This makes the problem considerably harder, requiring multiple handovers before final assembly.
The scenarios that we are looking at are summarized in \cref{tab:envs}.

For the simulation experiments, we use \qty{10}{\newton} insertion force, and no noise if not stated otherwise.
The wrench capabilities of the grippers are specified in \cref{tab:gripper_wrench_limits}.
We run the planner with the same settings in all environments, except for the grasp-set size: For the more complex mesh geometries of the Fabrica models, we use a larger fixed grasp set (64 compared to the standard 20) and disable grasp widening.
If not stated otherwise, we repeat all experiments 10 times.

\begin{table}[t]
    \centering
    \caption{Overview of the scenarios used to test our planner.}
      \resizebox{\linewidth}{!}{%
    \begin{tabular}{l|llccl}
        \toprule
        Env. & Robots & DoF & EE & \# parts & Dep. type\\
        \midrule
        Stool     & 4 UR5e &  24 & Mixed & 5 & DAG \\
        Chair     & 4 UR5e & 24 & Mixed &  8 & DAG\\
        Cross     & 4 UR5e &  24 & 2F & 7 & Sequence \\
        Cube      & 4 UR10e & 24 & Vac. & 8 & Sequence\\
        \midrule
        Fabrica (7 assemblies) & 4 UR5e & 24 & 2F & 5--9 & DAG\\
        \bottomrule
    \end{tabular}
    }
    \label{tab:envs}
\end{table}

\subsubsection{Computation times}
To the best of our knowledge, there is no other work taking the wrench constraints into account in the assembly planning problem, therefore comparing to an adequate baseline is difficult.
We thus compare against an approach that is similar to \cite{dogar2019multi}, i.e., we optimistically solve for the assembly grasps and poses, and then plan the required reorientation and handover actions.

\Cref{tab:comparison} shows the mean of the initial solution time (without preprocessing) per assembly and the mean cost of the solution at this time.
Compared to the Optimistic planner, our planner is usually faster, but (due to the greedy search) sometimes finds worse solutions.
The solution times for the assemblies from Fabrica are significantly higher than for the assemblies introduced here.
This is due to our current backend requiring a convex decomposition of the parts, leading to a more expensive collision check.
\Cref{tab:search_time_spent} gives details on where search time is spent:
The most significant differences are in the computation of the grasp database, the forward search, and the motion planning, all collision check heavy operations.
We can also see that the wrench LP and the heuristic computation take negligible time in most cases.

\begin{table*}[t]
  \centering
  \caption{Comparison of search times and costs for different algorithms. We report the mean and standard deviation of 10 runs.}
  \label{tab:comparison}
  \resizebox{\textwidth}{!}{%
  \begin{tabular}{ll|*{10}{S[table-format=3.2(2),separate-uncertainty=true,detect-weight=true,table-number-alignment=center,table-text-alignment=center]}}
\toprule
Solver & Metric & \multicolumn{1}{c}{Cross} & \multicolumn{1}{c}{Stool} & \multicolumn{1}{c}{Chair} & \multicolumn{1}{c}{F. Beam} & \multicolumn{1}{c}{F. Plumber block} & \multicolumn{1}{c}{F. Car} & \multicolumn{1}{c}{F. Gamepad} & \multicolumn{1}{c}{F. Cooling Manifold} & \multicolumn{1}{c}{F. Duct} & \multicolumn{1}{c}{F. Stool} \\
\midrule
\multirow{2}{*}{Optimistic} & $t_{\mathrm{init}}$ & 11.25 +- 0.15 & 14.42 +- 0.09 & \multicolumn{1}{c}{timeout} & 13.01 +- 0.03 & 9.73 +- 0.05 & 49.29 +- 0.23 & 115.13 +- 0.28 & 60.65 +- 0.08 & 51.46 +- 0.09 & \bfseries 20.59 +- 0.04 \\
 & cost & \bfseries 22 & \bfseries 20 & \multicolumn{1}{c}{timeout} & \bfseries 14 & \bfseries 14 & \bfseries 18 & \bfseries 19 & \bfseries 26 & \bfseries 25 & \bfseries 24 \\
\multirow{2}{*}{Ours} & $t_{\mathrm{init}}$ & 2.01 +- 0.02 & \bfseries 1.60 +- 0.02 & 8.93 +- 0.03 & \bfseries 9.73 +- 0.05 & 7.87 +- 0.02 & 19.71 +- 0.03 & 82.25 +- 0.13 & \bfseries 19.11 +- 0.11 & \bfseries 17.74 +- 0.26 & 27.70 +- 0.07 \\
 & cost & \bfseries 22 & 32 & 43 & 16 & \bfseries 14 & \bfseries 18 & 23 & 46 & \bfseries 25 & 30 \\
\midrule
\multirow{2}{*}{Flat lazy} & $t_{\mathrm{init}}$ & \multicolumn{1}{c}{OOM} & \multicolumn{1}{c}{OOM} & \multicolumn{1}{c}{OOM} & \multicolumn{1}{c}{OOM} & \multicolumn{1}{c}{OOM} & \multicolumn{1}{c}{OOM} & \multicolumn{1}{c}{OOM} & \multicolumn{1}{c}{OOM} & \multicolumn{1}{c}{OOM} & 45.12 +- 0.08 \\
 & cost & \multicolumn{1}{c}{OOM} & \multicolumn{1}{c}{OOM} & \multicolumn{1}{c}{OOM} & \multicolumn{1}{c}{OOM} & \multicolumn{1}{c}{OOM} & \multicolumn{1}{c}{OOM} & \multicolumn{1}{c}{OOM} & \multicolumn{1}{c}{OOM} & \multicolumn{1}{c}{OOM} & \bfseries 24 \\
\multirow{2}{*}{Milestones + counting} & $t_{\mathrm{init}}$ & \bfseries 1.69 +- 0.01 & 41.12 +- 0.30 & \bfseries 3.84 +- 0.02 & 10.26 +- 0.05 & \bfseries 7.48 +- 0.03 & 19.53 +- 0.04 & \bfseries 59.78 +- 0.13 & \multicolumn{1}{c}{timeout} & 47.32 +- 0.39 & 20.70 +- 0.29 \\
 & cost & \bfseries 22 & 24 & \bfseries 31 & \bfseries 14 & \bfseries 14 & \bfseries 18 & \bfseries 19 & \multicolumn{1}{c}{timeout} & \bfseries 25 & \bfseries 24 \\
\multirow{2}{*}{Ours, exact fills} & $t_{\mathrm{init}}$ & 1.98 +- 0.02 & 12.51 +- 0.04 & 5.53 +- 0.02 & 9.96 +- 0.03 & 7.87 +- 0.03 & \bfseries 19.37 +- 0.03 & 87.64 +- 0.17 & 20.19 +- 0.08 & 43.24 +- 0.29 & 31.55 +- 0.10 \\
 & cost & \bfseries 22 & \bfseries 20 & \bfseries 31 & \bfseries 14 & \bfseries 14 & \bfseries 18 & \bfseries 19 & \bfseries 26 & \bfseries 25 & \bfseries 24 \\
\multirow{2}{*}{Ours, fine projection} & $t_{\mathrm{init}}$ & 5.73 +- 0.02 & 2.02 +- 0.01 & 16.37 +- 0.06 & 10.34 +- 0.04 & 11.01 +- 0.04 & 22.10 +- 0.04 & 98.02 +- 0.25 & 50.02 +- 0.24 & 44.27 +- 0.42 & 28.84 +- 0.09 \\
 & cost & \bfseries 22 & 32 & 43 & 16 & \bfseries 14 & \bfseries 18 & 23 & 38 & \bfseries 25 & 30 \\
\bottomrule
  \end{tabular}%
  }
  \vspace{-3mm}
\end{table*}

\begin{table}[t]
  \centering
  \caption{Breakdown of where time is spent in \textsc{Wrap}. We report mean and standard deviation from 10 runs.}
  \label{tab:search_time_spent}
  \resizebox{\columnwidth}{!}{%
  \begin{tabular}{l|S[table-format=3.2(2),separate-uncertainty=true,table-number-alignment=center]|S[table-format=3.2(2),separate-uncertainty=true,table-number-alignment=center]S[table-format=3.2(2),separate-uncertainty=true,table-number-alignment=center]|S[table-format=3.2(2),separate-uncertainty=true,table-number-alignment=center]S[table-format=3.2(2),separate-uncertainty=true,table-number-alignment=center]|S[table-format=3.2(2),separate-uncertainty=true,table-number-alignment=center]}
\toprule
& \multicolumn{1}{c|}{} & \multicolumn{2}{c|}{Backwards} & \multicolumn{2}{c|}{Forward} & \multicolumn{1}{c}{Motion} \\
Object & \multicolumn{1}{c|}{$t_{\text{grasp}}$} & \multicolumn{1}{c}{$t_{\text{bwd search}}$} & \multicolumn{1}{c|}{$t_{\text{force LP}}$} & \multicolumn{1}{c}{$t_{\text{fwd search}}$} & \multicolumn{1}{c|}{$t_{\text{IK}}$} & \multicolumn{1}{c}{$t_{\text{MP init}}$} \\
\midrule
Cross & 0.48 +- 0.01 & 0.77 +- 0.02 & 0.19 +- 0.00 & 0.66 +- 0.01 & 0.40 +- 0.00 & 7.67 +- 0.57 \\
Stool & 2.43 +- 0.02 & 0.35 +- 0.00 & 0.05 +- 0.00 & 0.70 +- 0.02 & 0.50 +- 0.00 & 9.65 +- 0.46 \\
Chair & 11.81 +- 0.05 & 1.27 +- 0.01 & 0.40 +- 0.00 & 3.46 +- 0.03 & 3.80 +- 0.00 & 14.31 +- 0.68 \\
F. Beam & 8.99 +- 0.10 & 0.10 +- 0.01 & 0.05 +- 0.00 & 9.28 +- 0.05 & 0.30 +- 0.00 & 3.14 +- 0.20 \\
F. Plumb. block & 5.49 +- 0.03 & 0.64 +- 0.01 & 0.09 +- 0.00 & 5.24 +- 0.03 & 1.89 +- 0.03 & 44.36 +- 3.05 \\
F. Car & 6.93 +- 0.03 & 1.25 +- 0.00 & 0.03 +- 0.00 & 7.43 +- 0.03 & 11.00 +- 0.00 & 20.71 +- 1.91 \\
F. Gamepad & 49.87 +- 0.30 & 1.47 +- 0.01 & 0.11 +- 0.00 & 46.93 +- 0.11 & 33.74 +- 0.05 & 113.37 +- 14.25 \\
F. Cool. Mani. & 17.52 +- 0.04 & 1.79 +- 0.04 & 0.22 +- 0.00 & 14.98 +- 0.08 & 2.12 +- 0.04 & 38.51 +- 2.29 \\
F. Duct & 42.96 +- 0.13 & 1.76 +- 0.02 & 0.17 +- 0.01 & 13.39 +- 0.20 & 2.43 +- 0.05 & 125.36 +- 14.60 \\
F. Stool & 23.14 +- 0.05 & 2.71 +- 0.02 & 0.35 +- 0.00 & 20.64 +- 0.07 & 4.00 +- 0.00 & 65.54 +- 2.98 \\
\bottomrule
  \end{tabular}%
  }
\end{table}

\subsubsection{Scaling}
We analyze the scaling behavior of \textsc{Wrap} for more robots, and larger assemblies.
We test the scaling on the cross with robots with two-finger grippers, and on a 2x2x3 version of the cube assembly, using vacuum grippers.

\begin{figure}[t]
    \centering
    \begin{subfigure}[b]{0.9\linewidth}
        \centering
        \includegraphics[width=1\linewidth]{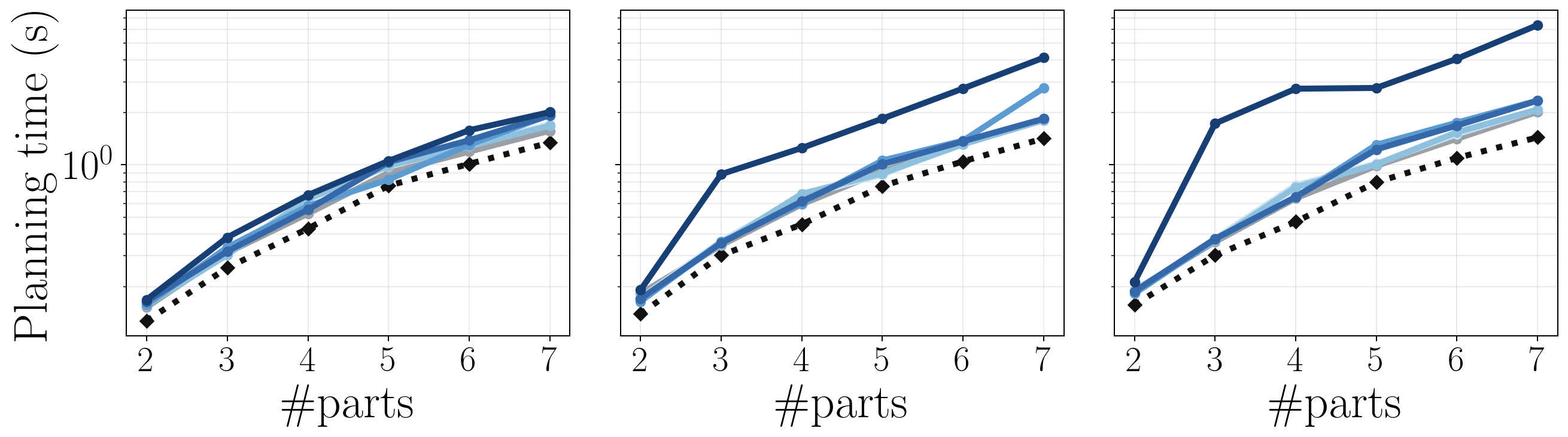}
        \caption{Cross, two-finger grippers.}
        \label{fig:scaling_no_wrench}
    \end{subfigure}
    \\
    \begin{subfigure}[b]{0.9\linewidth}
        \centering
        \includegraphics[width=1\linewidth]{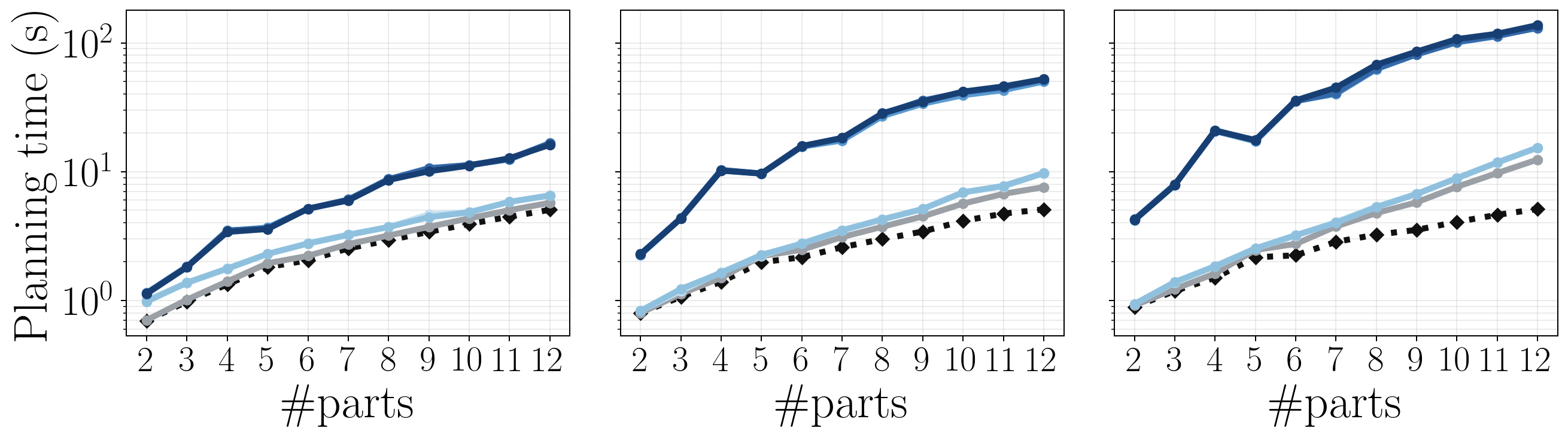}
        \caption{Cube, vacuum grippers.}
        \label{fig:scaling_wrench}
    \end{subfigure}
    \begin{subfigure}[b]{1.0\linewidth}%
        \centering
        \definecolor{processTopological}{HTML}{111111}
\definecolor{processOptimistic}{HTML}{9C2F2F}
\definecolor{processFailure}{HTML}{555555}
\definecolor{processLambdaZero}{HTML}{9AA0A6}
\definecolor{processLambdaPointZeroOne}{HTML}{C7E1F2}
\definecolor{processLambdaPointOne}{HTML}{8FC1DF}
\definecolor{processLambdaPointFive}{HTML}{5B9BD1}
\definecolor{processLambdaOne}{HTML}{3568A8}
\definecolor{processLambdaTwo}{HTML}{173F73}
\begin{tikzpicture}
  \begin{axis}[
    width=\columnwidth,
    height=0.4\columnwidth,
    hide axis,
    xmin=0, xmax=10, ymin=0, ymax=10,
    legend style={
      draw=none,
      anchor=center,
      legend cell align=left,
      legend columns=4,
      at={(axis cs:5,5)},
      font=\scriptsize,
      /tikz/every even column/.append style={column sep=0.35em}
    }
  ]
    \addlegendimage{color=processTopological, dotted, line width=1.0pt, mark=diamond*, mark size=1.5pt}
    \addlegendentry{w/o wrench}
    \addlegendimage{color=processLambdaZero, solid, line width=1.0pt, mark=*, mark size=1.5pt}
    \addlegendentry{$\lambda=0$}
    \addlegendimage{color=processLambdaPointZeroOne, solid, line width=1.0pt, mark=*, mark size=1.5pt}
    \addlegendentry{$\lambda=0.01$}
    \addlegendimage{color=processLambdaPointOne, solid, line width=1.0pt, mark=*, mark size=1.5pt}
    \addlegendentry{$\lambda=0.1$}
    \addlegendimage{color=processLambdaPointFive, solid, line width=1.0pt, mark=*, mark size=1.5pt}
    \addlegendentry{$\lambda=0.5$}
    \addlegendimage{color=processLambdaOne, solid, line width=1.0pt, mark=*, mark size=1.5pt}
    \addlegendentry{$\lambda=1$}
    \addlegendimage{color=processLambdaTwo, solid, line width=1.0pt, mark=*, mark size=1.5pt}
    \addlegendentry{$\lambda=2$}
  \end{axis}
\end{tikzpicture}
    \end{subfigure}
    \caption{Scaling comparison for two objects and different wrench models for (\textbf{left}) two, (\textbf{middle}) three, and (\textbf{right}) four robots. $\lambda$ is a scaling factor on the required assembly force.}
    \label{fig:scaling}
\end{figure}

\Cref{fig:scaling} shows roughly linear scaling in the number of parts in compute time, and also shows that the computation time increases significantly when adding more robots.
This increase in computation time when adding more robots is caused by the larger search space: there are many more possibilities for valid pick assignments and handovers.

\paragraph*{Wrench/no-wrench} 
\Cref{fig:scaling} also shows planning times when not including any forces (what most other assembly plans do), including gravity only, and scaling the mating forces with $\lambda$.
Not requiring the grasps to satisfy wrench constraints, or only satisfying gravity, the search times are significantly lower (and the plans are shorter), as there are fewer constraints needing to be satisfied.

\subsubsection{Ablations}
\label{sssec:ablations}
We show a series of ablations to justify the choices in the planner.
Particularly, we show in \cref{tab:comparison}:
\begin{itemize}
    \item \emph{Flat lazy} searches over the complete task-state, i.e., not decomposing into subproblems according to the assembly dependency graph.
    \item \emph{Milestones + counting} uses the same milestones as \textsc{Wrap}, but replaces the backward search with a simple heuristic that counts the minimum number of pick, assembly, and release actions still required.
    \item \emph{Exact fills} replaces the greedy search with a weight-one lazy A* search.
    \item \emph{Fine projection} replaces the backward abstraction with a finer projection.
\end{itemize}

On average, \textsc{Wrap} obtains the initial solution fastest, but in many cases the simple counting heuristic and the optimistic search are competitive.
\emph{Flat lazy} stops due to running out of memory for all problems due to the large search space.

We also show the solutions cost of the plan, i.e., the number of actions that are taken in \cref{tab:comparison}.
The optimal forward search in the local fills leads to better solution than the greedy search in \textsc{Wrap}, but at the cost of being slower.

\subsubsection{Convergence and Optimality}
\begin{figure}[t]
    \centering
    \includegraphics[width=0.3\linewidth]{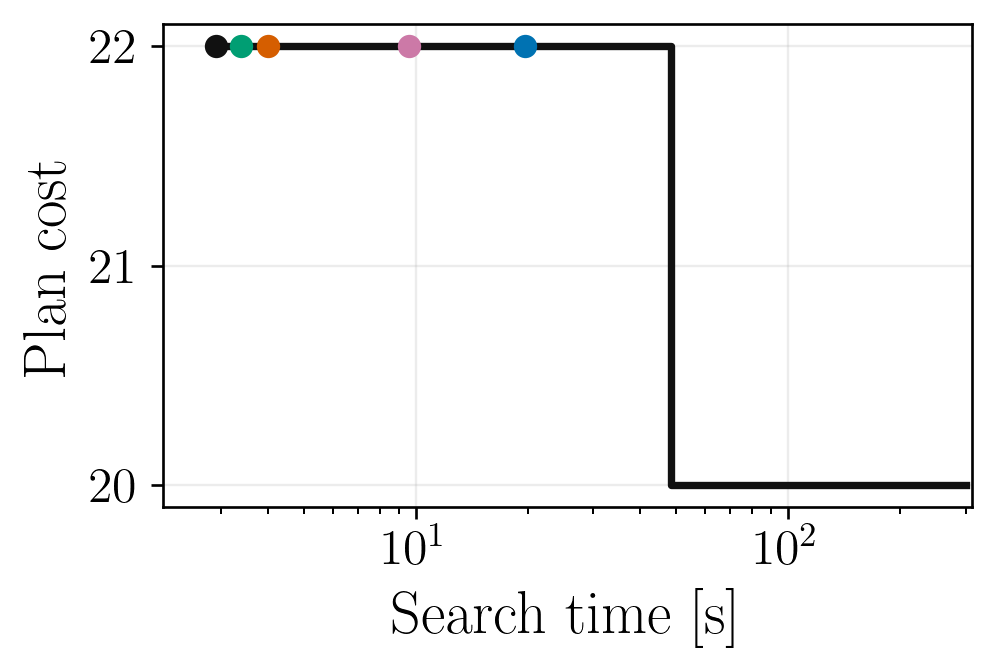}
    \includegraphics[width=0.3\linewidth]{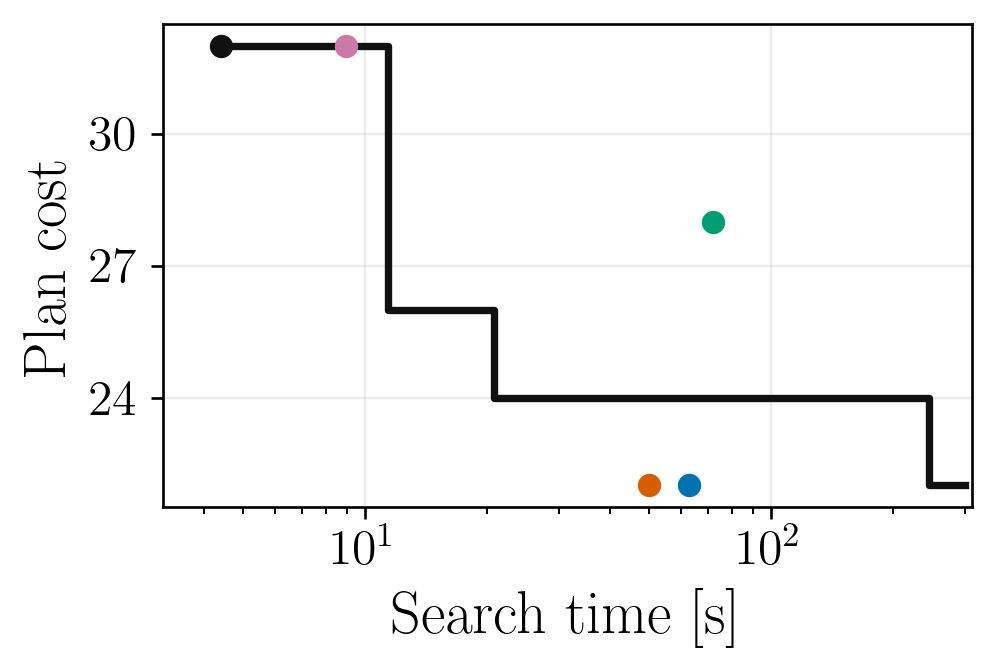}
    \includegraphics[width=0.3\linewidth]{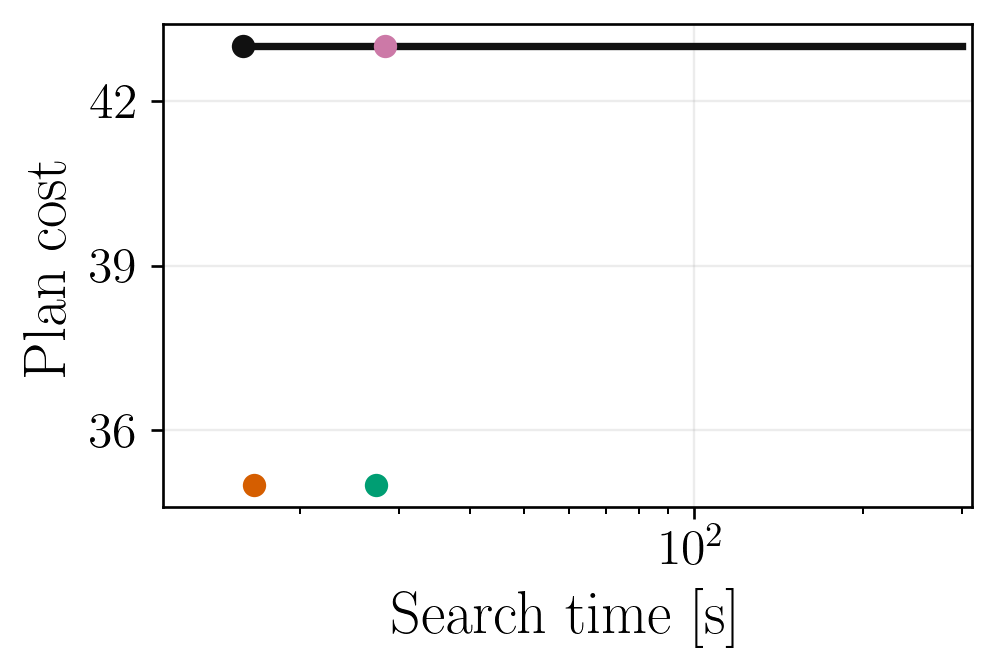}
        \begin{subfigure}[b]{1.0\linewidth}%
        \centering
        \definecolor{convergenceOurs}{HTML}{111111}
\definecolor{convergenceOptimistic}{HTML}{0072B2}
\definecolor{convergenceFlatLazy}{HTML}{999999}
\definecolor{convergenceCounting}{HTML}{009E73}
\definecolor{convergenceExactFills}{HTML}{D55E00}
\definecolor{convergenceFineProjection}{HTML}{CC79A7}
\begin{tikzpicture}
  \begin{axis}[
    width=\columnwidth,
    height=0.19\columnwidth,
    hide axis,
    xmin=0, xmax=10, ymin=0, ymax=10,
    legend style={
      draw=none, anchor=center, legend cell align=left,
      legend columns=6, at={(axis cs:5,5)}, font=\scriptsize,
      /tikz/every even column/.append style={column sep=0.45em}
    }
  ]
    \addlegendimage{color=convergenceOurs, solid, line width=1.0pt, mark=*, mark size=1.35pt}
    \addlegendentry{Ours}
    \addlegendimage{color=convergenceOptimistic, only marks, mark=*, mark size=1.35pt}
    \addlegendentry{Optimistic}
    \addlegendimage{color=convergenceFlatLazy, only marks, mark=*, mark size=1.35pt}
    \addlegendentry{Flat lazy}
    \addlegendimage{color=convergenceCounting, only marks, mark=*, mark size=1.35pt}
    \addlegendentry{Counting}
    \addlegendimage{color=convergenceExactFills, only marks, mark=*, mark size=1.35pt}
    \addlegendentry{Exact fills}
    \addlegendimage{color=convergenceFineProjection, only marks, mark=*, mark size=1.35pt}
    \addlegendentry{Fine projection}
  \end{axis}
\end{tikzpicture}
        \vspace{-5mm}
    \end{subfigure}
    \caption{Cost convergence plot for \textbf{(left)} Cross, \textbf{(middle)} Stool and \textbf{(right)} Chair.}
    \label{fig:search_conv}
\end{figure}

\textsc{Wrap} continues to search for a better task plan and decreases the cost until the timeout is reached.
We show a cost convergence plot (\cref{fig:search_conv}) for three assemblies, namely 'Cross', 'Stool', and 'Chair'.
While \textsc{Wrap} usually finds a solution earlier than the other planners, the initial cost is usually worse.
If given enough time, our planner often converges to the same or better solution cost when we keep optimizing the task plan.
A counterexample to this is `Chair', which consists of many parts and submodules.
There's two effects: (1) while we can optimize within the chosen assembly sequence, we might not be able to find the optimal solution, and (2) when we improve the milestone fills, we constrain ourselves to the previously chosen end-state.
If that end-state is suboptimal, we can only escape this when filling in longer horizons.

\subsubsection{Generalizations}
\begin{figure*}[t]
    \centering
    \includegraphics[width=0.19\linewidth,trim=0 1.5cm 0 2.5cm,clip]{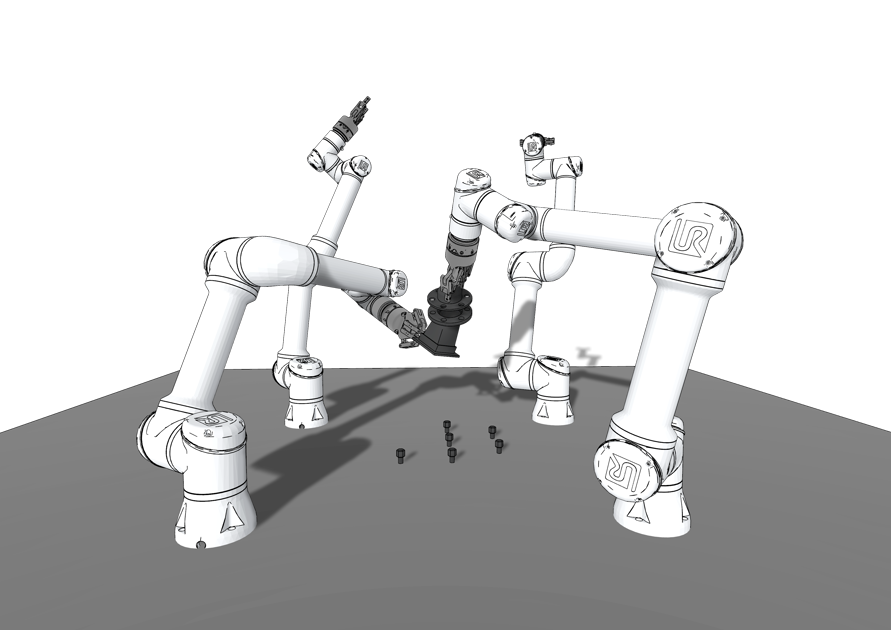}\hfill
    \includegraphics[width=0.19\linewidth,trim=0 1.5cm 0 2.5cm,clip]{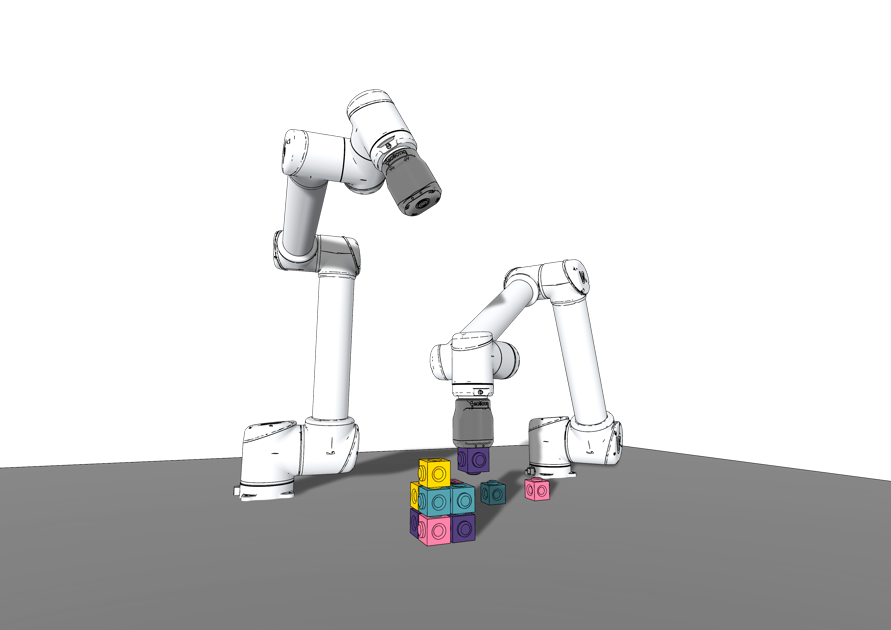}\hfill
    \includegraphics[width=0.19\linewidth,trim=0 1.5cm 0 2.5cm,clip]{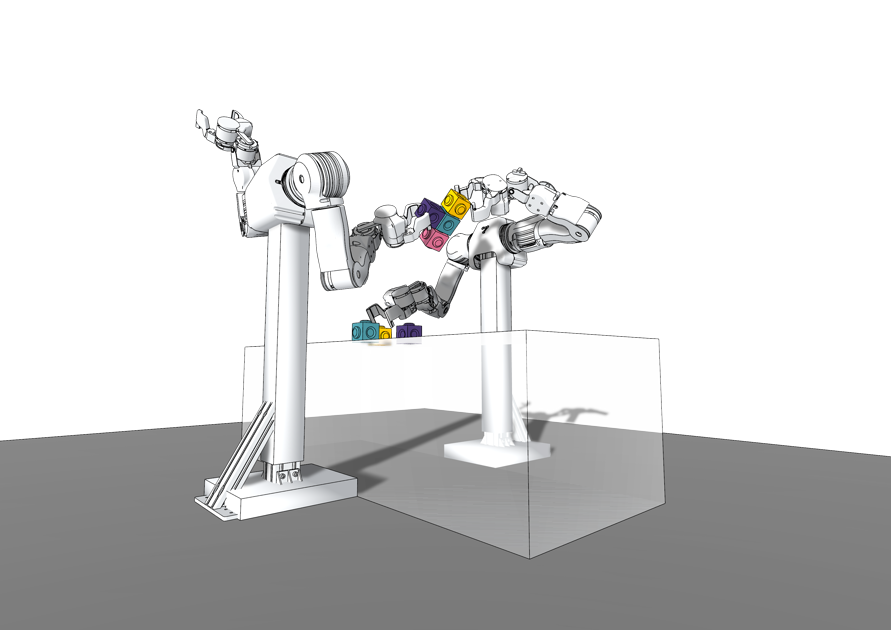}\hfill
    \includegraphics[width=0.19\linewidth,trim=0 1.5cm 0 2.5cm,clip]{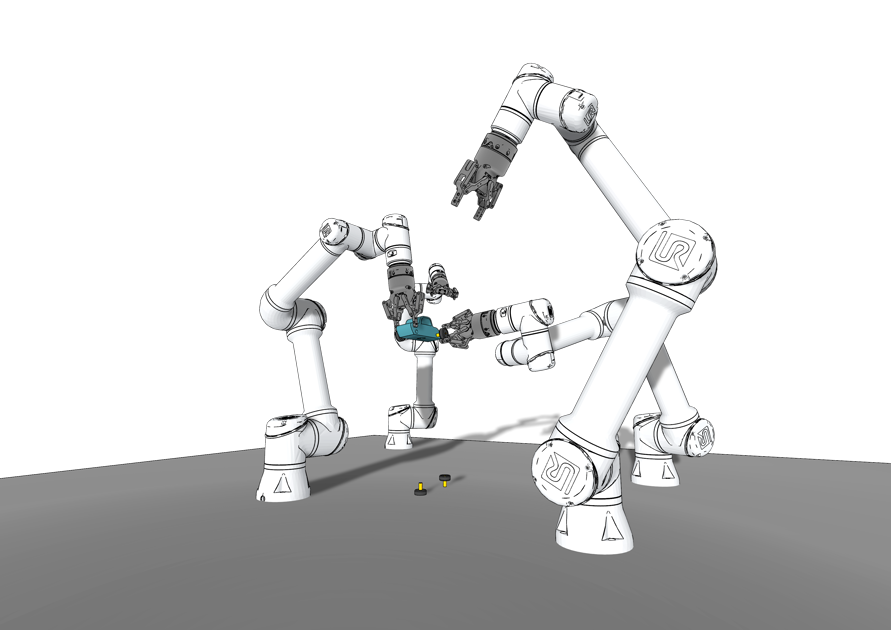}\hfill
    \includegraphics[width=0.19\linewidth,trim=0 1.5cm 0 2.5cm,clip]{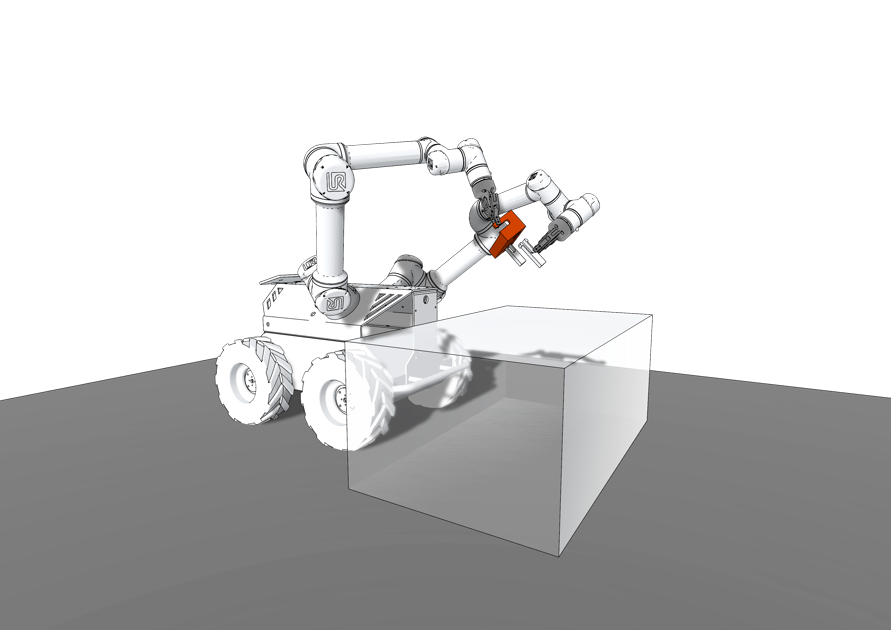}\hfill
    \caption{Snapshots from simulation, from left to right: Duct, 2x2x3 Cube, Cross with OpenArm, Car, Half-Stool with Husky.}
    \label{fig:sim_img}
\end{figure*}

We demonstrate two extensions with no changes to \textsc{Wrap}: the Cross assembly using four seven-DoF OpenArm manipulators (\cref{fig:sim_img}), and a five-part stool whose four legs are attached using screw motions rather than linear insertions in the supplementary material.

\subsection{Simulation}
We run the paths that we compute in a physics simulator (MuJoCo) to ensure feasibility under the assembly wrenches and the robot dynamics. 
Since real-to-sim is a difficult problem to solve in itself (i.e., matching friction behavior, and contact-rich insertion behavior from reality to the simulation), we rigidly attach the parts to the grippers and manually check the wrench at the interface and apply the mating wrenches as external wrenches that are applied to the bodies.

We compute a path with the wrench gates in the search, one with only gravity, and one without any wrench checks. 
\Cref{tab:wrench_sim} reports the utilization of the wrench limits separately for the vacuum grippers and the two-finger grippers in the scene, and we report the utilization during transport and during insertion separately.

Plans can fail with peak utilization lower than 100 with other failure modes than fail-during-insertion: Both approaches that do not take the assembly wrench into account do not manage to complete an assembly.
The most common failures for `Cross' and `Cube' are attempting to assemble objects without support on the table, and the object sliding away, and for `Chair' using the vacuum gripper in a direction that does not support a force.
On the cube example, \textsc{Wrap} fails once due to object misalignment when assembling.

\begin{table}[t]
\centering
\caption{Simulation success of plans computed with different support models.
We report success and mean peak support utilization over 10 runs for the successful part of the trajectory.
Utilization is split by execution phase and gripper type; bold values exceed
the limit. -- denotes an absent gripper type.}
\begingroup
\footnotesize
\setlength{\tabcolsep}{1.5pt}
\begin{tabular*}{\columnwidth}{@{\extracolsep{\fill}}llrlccccc@{}}
\toprule
Env. & Model & Succ. & Tasks & Fail. &
\multicolumn{2}{c}{Transport util.} &
\multicolumn{2}{c}{Insertion util.} \\
\cmidrule(lr){6-7}\cmidrule(lr){8-9}
& & [\%] & & & 2F & Vac. & 2F & Vac. \\
\midrule
\multirow{3}{*}{Cross}
& w/o wrench  & 0   & 11/26   & no support (10) & 10 & -- & 18 & -- \\
& Gravity & 0   & 11/26   & no support (10) & 10 & -- & 18 & -- \\
& Full    & 100 & 22/22   & --           & 25   & -- & 21   & -- \\
\midrule
\multirow{3}{*}{Cube}
& w/o wrench  & 0  & 1.4/21  & no support (10) & -- & 8  & -- & 34 \\
& Gravity & 0  & 1.4/21  & no support (10) & -- & 8  & -- & 34 \\
& Full    & 90 & 41.4/44 & seat (1)     & -- & 47 & -- & 28 \\
\midrule
\multirow{3}{*}{Chair}
& w/o wrench  & 0   & 2/30  & vac. (10) & 0 & 34 & 0 & $\mathbf{104}$ \\
& Gravity & 0   & 2/30  & vac. (10) & 0 & 36 & 0 & $\mathbf{103}$ \\
& Full    & 100 & 36/36 & --        & 18  & 95   & 19  & 89 \\
\bottomrule
\end{tabular*}
\endgroup
\vspace{-2mm}
\label{tab:wrench_sim}
\end{table}

\subsection{Real world execution}
We show real robot execution on the example of a version of the stool with two UR5e robots mounted on a stationary husky base, with snapshots in \cref{fig:robot_eval}, and a video in the supplementary material.
We cut the stool in half for the demo: The plan then consists of 7 actions, containing a handover, and the two assembly steps inserting the legs into the base.
We track the reference path with a joint space velocity controller, and we run a compliant controller for the insertion.
The compliant controller was enough to successfully insert the legs in the stool, since we know the initial positions of the parts due to an accurate calibration of the system, and, e.g., vision based recovery was not needed.

\begin{figure}[t]
    \centering
    \includegraphics[width=0.9\linewidth]{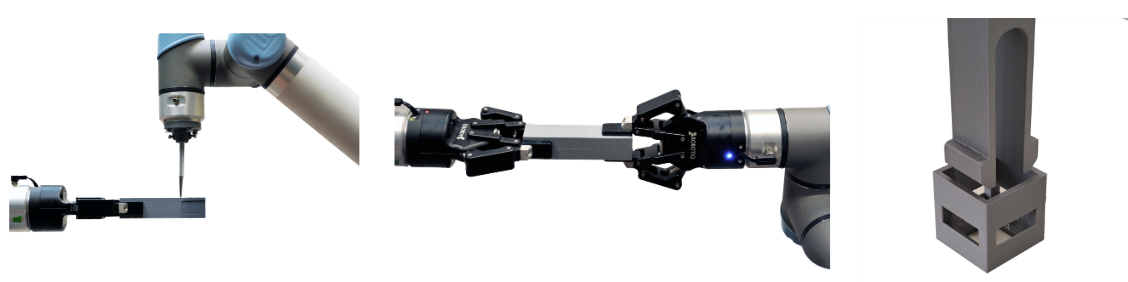}
    \caption{Robot calibration (left and middle) and closeup of the snap fit (right) that we use in the stool model.}
    \label{fig:gripper_calibration}
\end{figure}

\paragraph*{Gripper wrench and insertion wrench calibration}
The set of wrenches $\mathcal{W}_r$ that the gripper can transmit to the object without slippage is dependent on the gripper-object interface.
We measure the range by grasping an object of the correct material, and applying an external force until the object moves, either through slipping, or through the gripper letting go.
\cref{fig:gripper_calibration} shows the setup for the measurement of the slippage-limits, and \cref{tab:gripper_wrench_limits} lists the manufacturer limit of the gripper, and the calibrated limits at which slippage happens.
To be conservative in what forces and torques we apply, we take the minimum of both values.

We also measure the forces that happen during the mating of the parts that we execute on the real robot system.
The maximum forces we measure during multiple test insertions were \qty{17.32}{\newton} axial force, and \qty{5.38}{\newton} transverse force.
The maximum torques were \qtylist{0.70; 1.34; 2.08}{\newton\meter}.
With a safety factor, we therefore set the nominal force that we plan with for the real execution to \qty{25}{\newton}.

\begin{table}[t]
    \centering
    \caption{Wrench limits used during assembly and transport. Two-finger
values are the minimum of the manufacturer limit and the
measured slippage load (in parentheses). The same limits apply to
assembly and transport. For the vacuum gripper, only a $z$-force is
allowed during assembly, since deflection otherwise occurs.}
    \footnotesize
    \begin{tabular}{l >{\centering\arraybackslash}p{3.5em}@{\,\,}>{\centering\arraybackslash}p{2.4em} c c}
        \toprule
        & \multicolumn{2}{c}{Two-finger gripper}
        & \multicolumn{2}{c}{Vacuum gripper} \\
        \cmidrule(lr){2-3}
        \cmidrule(lr){4-5}
        {Parameter}
        & \multicolumn{2}{c}{\shortstack{Adopted\\(measured)}}
        & \shortstack{Assembly\\measured}
        & \shortstack{Transport\\measured} \\
        \midrule
        $F_x$ [\si{\newton}]       & \hspace{8pt}50 & (95.3)  & 0 & 67.4 \\
        $F_y$ [\si{\newton}]       & \hspace{8pt}39.7   & (39.7)  & 0 & 67.4 \\
        $F_z$ [\si{\newton}]       & \hspace{8pt}50 & (80.4)  & 110.0   & 110.0 \\
        $M_x$ [\si{\newton\meter}]  & \hspace{8pt}4.7   & (4.7)   & 0 & 5.0 \\
        $M_y$ [\si{\newton\meter}]  & \hspace{8pt}4.5   & (4.5)   & 0 & 5.0 \\
        $M_z$ [\si{\newton\meter}]  & \hspace{8pt}3  & (9.8)  & 0 & 2.1 \\
        \bottomrule
    \end{tabular}
    \label{tab:gripper_wrench_limits}
\end{table}

\begin{figure*}[t]
    \centering
    \includegraphics[width=0.16\linewidth]{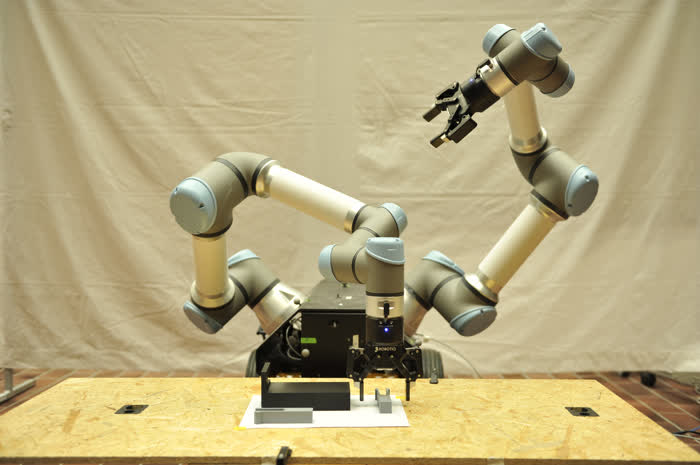}\hfill
    \includegraphics[width=0.16\linewidth]{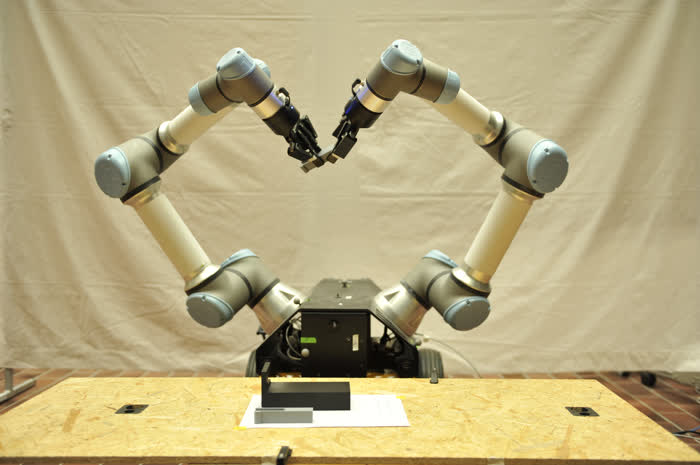}\hfill
    \includegraphics[width=0.16\linewidth]{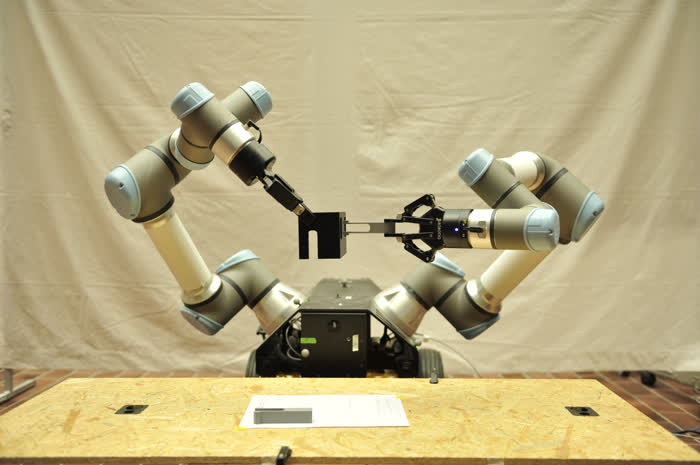}\hfill
    \includegraphics[width=0.16\linewidth]{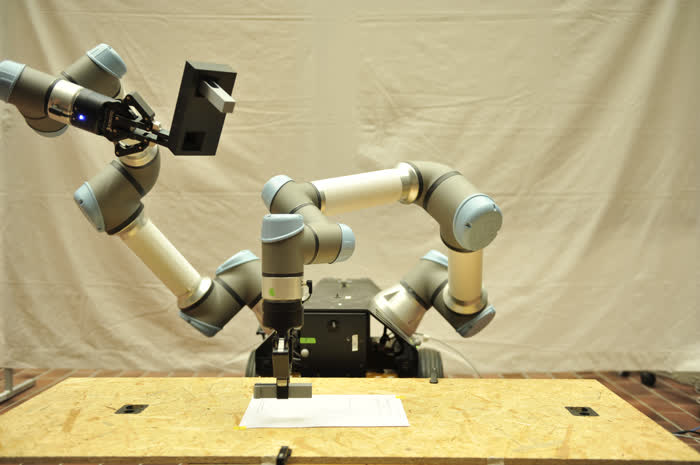}\hfill
    \includegraphics[width=0.16\linewidth]{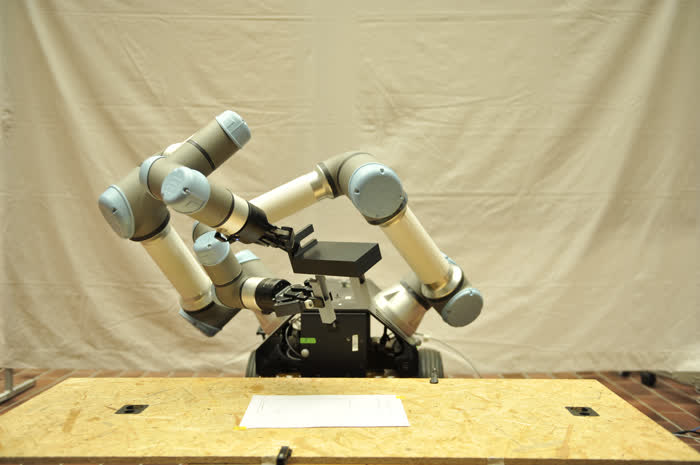}\hfill
    \includegraphics[width=0.16\linewidth]{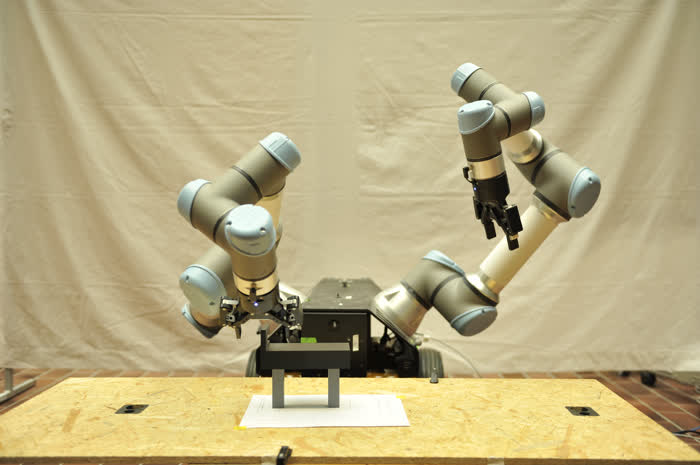}
    \caption{Sequence from the real robot execution of the stool assembly with two UR5e arms on a stationary Husky base. From left to right: initial configuration, handing over the first leg, grasping the seat, inserting a leg into the seat held in mid-air, grasping the second leg, and insertion, and placing the assembled stool on the table.}
    \label{fig:robot_eval}
    \vspace{-6mm}
\end{figure*}

\section{Discussion \& Limitations}

The approach we present here is based on a classical search, which means that eventually, in order to certify infeasibility, all possible actions and parameterizations have to be tested, which is not scaleable for large assemblies and many robots.
This effect is visible in the scaling experiments, where the performance is influenced both by the number of robots, and the number of parts.
In the settings where we solved the same assembly with a different number of robots, ideally the runtime would not increase from involving more robots if it is possible to solve with fewer robots.
One could deal with this by focusing on subsets of the whole state space, i.e., not considering all robots from the beginning on.

While we showed the execution of the plans on real robots, closed loop execution of plans is left as future work.
The controllers currently absorb some inaccuracies, but a replanning approach is required to deal with larger execution failures.
The approach is currently limited to prehensile manipulation, and actions that have deterministic effects, currently preventing, e.g., pushing objects away to reach a part.
We intend to combine this task planning approach with learned policies, enabling both non-prehensile manipulation, and, e.g., dexterous manipulation for in-hand reorientation instead of requiring multiple handovers.
In the real robot execution, the rigidity assumption per part does not necessarily hold anymore: Depending on the grasp, some parts bent under load: internal stress should be taken into account in the task planning, in addition to the external forces.

\section{Conclusion}
We presented \textsc{Wrap}, a method to compute robot task assignments and sequences in order to compute a multi-robot action plan from a given assembly dependency graph for a multi-part assembly.
\textsc{Wrap} takes the forces that happen during assembly steps into account, does not require fixtures, and computes asynchronous motion plans that are executable on a real multi-robot system.

{
\footnotesize
\section{Acknowledgment}
This work was part of an Innovation project supported by Innosuisse.
YH was supported by the ETH postdoc fellowship and the SNSF Ambizione Grant (Grant No. 223384)
We used coding agents to rewrite initial implementations, including Claude Opus and Fable 5, GPT 5.6 Sol, GLM 5.3, Deepseek V4.
}

\bibliography{bib}

@String(rss = "Proc{.} of Robotics: Science and Systems (R:SS)")

@String(icra = "Proc{.} of the IEEE Int{.} Conf{.} on Robotics and Automation (ICRA)")

@String(iros = "Proc{.} of the IEEE/RSJ Int{.} Conf{.} on Intelligent Robots and Systems (IROS)")

@String(tro = "{IEEE} Trans. on Robotics")

@String(auto = "Autonomous Robots")

@String(ijrr = "International Journal of Robotics Research")

@String(wafr = "World Symposium on the Algorithmic Foundations of Robotics (WAFR)")

@String(ral = "{IEEE} Robotics and Automation Letters")

@String(ral = "{IEEE} Robot. and Automat. Lett. (R-AL)")

@String(ras = "{IEEE} Robotics and Autonomous Systems")

@String(corl = "Conf. on Robot Learning ({CoRL})")

@String(icml = "Int. Conf. on Machine Learning ({ICML})")

@article{dogar2019multi,
  title={Multi-robot grasp planning for sequential assembly operations},
  author={Dogar, Mehmet and Spielberg, Andrew and Baker, Stuart and Rus, Daniela},
  journal=auto,
  volume={43},
  number={3},
  pages={649--664},
  year={2019},
  publisher={Springer}
}

@article{tian2022assemble,
    title={Assemble Them All: Physics-Based Planning for Generalizable Assembly by Disassembly},
    author={Tian, Yunsheng and Xu, Jie and Li, Yichen and Luo, Jieliang and Sueda, Shinjiro and Li, Hui and Willis, Karl D.D. and Matusik, Wojciech},
    journal={ACM Trans. Graph.},
    year={2022},
    publisher={ACM}
}

@inproceedings{huang2025apexmr,
              title = {APEX-MR: Multi-Robot Asynchronous Planning and Execution for Cooperative Assembly},
              author = {Huang, Philip and Liu, Ruixuan and Aggarwal, Shobhit and Liu, Changliu and Li, Jiaoyang},
              year = {2025},
              booktitle = rss,
            }

@inproceedings{tian2025fabrica,
  title={Fabrica: Dual-Arm Assembly of General Multi-Part Objects via Integrated Planning and Learning},
  author={Yunsheng Tian and Joshua Jacob and Yijiang Huang and Jialiang Zhao and Edward Li Gu and Pingchuan Ma and Annan Zhang and Farhad Javid and Branden Romero and Sachin Chitta and Shinjiro Sueda and Hui Li and Wojciech Matusik},
  booktitle=corl,
  year={2025},
}

@article{22-hartmann-TRO,
  title = {Long-Horizon Multi-Robot Rearrangement Planning for
  		  Construction Assembly},
  author = {Hartmann, Valentin N. and Orthey, Andreas and Driess, Danny and Oguz, Ozgur S. and Toussaint, Marc},
  journal = tro,
  issn = {1552-3098},
  year = {2022},
  arxiv_pdf = {2106.02489},
  youtube = {GqhouvL5dig},
  web = {https://vhartmann.com/multi-robot/},
  doi = {10.1109/TRO.2022.3198020}
}

@article{hargus2026coordinated,
  title={Coordinated Multi-Robot Disassembly for Makespan Optimization of Large-Scale Assemblies},
  author={Hargus, Niklas and Orthey, Andreas and Toussaint, Marc},
  journal={arXiv preprint arXiv:2608.05830},
  year={2026}
}

@article{bayraktar2026peel,
  title={PEEL: Parallel Extraction for Long-Horizon Disassembly Planning via Scale-Invariant Sampling},
  author={Bayraktar, Servet B and Orthey, Andreas and Kingston, Zachary and Toussaint, Marc},
  journal={arXiv preprint arXiv:2608.08773},
  year={2026}
}

@inproceedings{hartmann2025benchmark,
  title={Sampling-Based Multi-Modal Multi-Robot Multi-Goal Path Planning},
  author={Hartmann, Valentin N and Heinle, Tirza and Huang, Yijiang and Coros, Stelian},
  booktitle = wafr,
  year = {2026},
  date = {2026-06-15},
}

@article{chen2022coop, 
  title={Cooperative Task and Motion Planning for Multi-Arm Assembly Systems}, 
  author={Chen, Jingkai and Li, Jiaoyang and Huang, Yijiang and Garrett, Caelan and Sun, Dawei and Fan, Chuchu and Hofmann, Andreas and Mueller, Caitlin and Koenig, Sven and Williams, Brian C.}, 
  year={2022}, month={Mar},
  url={http://arxiv.org/abs/2203.02475}, 
  DOI={10.48550/arXiv.2203.02475}, 
}

@inproceedings{tian2024asap,
  title={{Asap: Automated sequence planning for complex robotic assembly with physical feasibility}},
  author={Tian, Yunsheng and Willis, Karl DD and Al Omari, Bassel and Luo, Jieliang and Ma, Pingchuan and Li, Yichen and Javid, Farhad and Gu, Edward and Jacob, Joshua and Sueda, Shinjiro and others},
  booktitle=icra,
  year={2024},
  organization={IEEE}
}

@inproceedings{nagele2020legobot,
  title={Legobot: Automated planning for coordinated multi-robot assembly of lego structures},
  author={N{\"a}gele, Ludwig and Hoffmann, Alwin and Schierl, Andreas and Reif, Wolfgang},
  booktitle=iros,
  pages={9088--9095},
  year={2020},
  organization={IEEE}
}

@article{suarez2018can,
  title={Can robots assemble an IKEA chair?},
  author={Su{\'a}rez-Ruiz, Francisco and Zhou, Xian and Pham, Quang-Cuong},
  journal={Science Robotics},
  volume={3},
  number={17},
  pages={eaat6385},
  year={2018},
  publisher={American Association for the Advancement of Science}
}

@article{20-toussaint-RAL,
  title = {Describing Physics For Physical Reasoning: Force-based
  		  Sequential Manipulation Planning},
  author = {Toussaint, Marc and Ha, Jung-Su and Driess, Danny},
  journal = {IEEE Robotics and Automation Letters},
  booktitle = iros,
  year = {2020},
  youtube = {tVFkKIIODaM},
  arxiv = {2002.12780},
  doi = {10.1109/LRA.2020.3010462}
}

@article{fang2023anygrasp,
  title={AnyGrasp: Robust and Efficient Grasp Perception in Spatial and Temporal Domains},
  author = {Fang, Hao-Shu and Wang, Chenxi and Fang, Hongjie and Gou, Minghao and Liu, Jirong and Yan, Hengxu and Liu, Wenhai and Xie, Yichen and Lu, Cewu},
  journal=tro,
  year={2023}
}

@Inbook{Pisinger2019,
author="Pisinger, David
and Ropke, Stefan",
editor="Gendreau, Michel
and Potvin, Jean-Yves",
title="Large Neighborhood Search",
bookTitle="Handbook of Metaheuristics",
year="2019",
publisher="Springer International Publishing",
address="Cham",
pages="99--127",
}

@inproceedings{toussaint2024effort,
  title={Effort level search in infinite completion trees with application to task-and-motion planning},
  author={Toussaint, Marc and Ortiz-Haro, Joaquim and Hartmann, Valentin N and Karpas, Erez and H{\"o}nig, Wolfgang},
  booktitle=icra,
  pages={14902--14908},
  year={2024},
  organization={IEEE}
}

@article{halperin2000general,
  title={A general framework for assembly planning: The motion space approach},
  author={Halperin, Dan and Latombe, Jean-Claude and Wilson, Randall H.},
  journal={Algorithmica},
  volume={26},
  number={3},
  pages={577--601},
  year={2000}
}

@inproceedings{sundaram2001disassembly,
  title={Disassembly sequencing using a motion planning approach},
  author={Sundaram, Sujay and Remmler, Ian and Amato, Nancy M.},
  booktitle=icra,
  volume={2},
  pages={1475--1480},
  year={2001}
}

@article{jimenez2013survey,
  title={Survey on assembly sequencing: a combinatorial and geometrical perspective},
  author={Jim{\'e}nez, Pablo},
  journal={Journal of Intelligent Manufacturing},
  volume={24},
  number={2},
  pages={235--250},
  year={2013}
}

@article{bahubalendruni2016review,
  title={A review on assembly sequence generation and its automation},
  author={Bahubalendruni, M. V. A. Raju and Biswal, Bibhuti Bhusan},
  journal={Proceedings of the Institution of Mechanical Engineers, Part C: Journal of Mechanical Engineering Science},
  volume={230},
  number={5},
  pages={824--838},
  year={2016}
}

@article{rodriguez2019iteratively,
  title={Iteratively refined feasibility checks in robotic assembly sequence planning},
  author={Rodr{\'\i}guez, Ismael and Nottensteiner, Korbinian and Leidner, Daniel and Ka{\ss}ecker, Michael and Stulp, Freek and Albu-Sch{\"a}ffer, Alin},
  journal=ral,
  year={2019}
}

@inproceedings{zhu2024multilevel,
  title={Multi-level reasoning for robotic assembly: From sequence inference to contact selection},
  author={Zhu, Xinghao and Jha, Devesh K. and Romeres, Diego and Sun, Lingfeng and Tomizuka, Masayoshi and Cherian, Anoop},
  booktitle=icra,
  pages={816--823},
  year={2024}
}

@article{marvel2018multirobot,
  title={Multi-robot assembly strategies and metrics},
  author={Marvel, Jeremy A. and Bostelman, Roger and Falco, Joe},
  journal={ACM Computing Surveys},
  volume={51},
  year={2018},
  doi={10.1145/3150225}
}

@inproceedings{moriyama2019dualarm,
  title={Dual-arm assembly planning considering gravitational constraints},
  author={Moriyama, Ryota and Wan, Weiwei and Harada, Kensuke},
  booktitle=iros,
  pages={5566--5572},
  year={2019},
  doi={10.1109/IROS40897.2019.8967883}
}

@article{mattikalli1995gravitational,
  title={Gravitational stability of frictionless assemblies},
  author={Mattikalli, Raju and Baraff, David and Khosla, Pradeep and Repetto, Bruno},
  journal=tro,
  year={1995}
}

@article{mattikalli1996stable,
  title={Finding all stable orientations of assemblies with friction},
  author={Mattikalli, Raju and Baraff, David and Khosla, Pradeep},
  journal=tro,
  volume={12},
  pages={290--301},
  year={1996}
}

@article{rakshit2015influence,
  title={The influence of motion paths and assembly sequences on the stability of assemblies},
  author={Rakshit, Sourav and Akella, Srinivas},
  journal={IEEE Trans. on Automation Science and Engineering},
  volume={12},
  number={2},
  pages={615--627},
  year={2015}
}

@article{holladay2024robust,
  title={Robust planning for multi-stage forceful manipulation},
  author={Holladay, Rachel and Lozano-P{\'e}rez, Tom{\'a}s and Rodriguez, Alberto},
  journal=ijrr,
  volume={43},
  number={3},
  pages={330--353},
  year={2024},
  doi={10.1177/02783649231198560}
}

@inproceedings{tang2024automate,
  title={{AutoMate}: Specialist and generalist assembly policies over diverse geometries},
  author={Tang, Bingjie and Akinola, Iretiayo and Xu, Jie and Wen, Bowen and Handa, Ankur and Van Wyk, Karl and Fox, Dieter and Sukhatme, Gaurav S. and Ramos, Fabio and Narang, Yashraj},
  booktitle=rss,
  year={2024}
}

@article{zhang2020cspace,
  title={C-space tunnel discovery for puzzle path planning},
  author={Zhang, Xinya and Belfer, Robert and Kry, Paul G. and Vouga, Etienne},
  journal={ACM Trans. on Graphics},
  year={2020}
}

@inproceedings{funk2022learn2assemble,
  title={Learn2Assemble with structured representations and search for robotic architectural construction},
  author={Funk, Niklas and Chalvatzaki, Georgia and Belousov, Boris and Peters, Jan},
  booktitle=corl,
  year={2022}
}

@inproceedings{ghasemipour2022blocks,
  title={Blocks assemble! {Learning} to assemble with large-scale structured reinforcement learning},
  author={Ghasemipour, Seyed Kamyar Seyed and Kataoka, Satoshi and David, Byron and Freeman, Daniel and Gu, Shixiang Shane and Mordatch, Igor},
  booktitle=icml,
  pages={7435--7469},
  year={2022}
}

@article{huang2021robotic,
  title={Robotic additive construction of bar structures: Unified sequence and motion planning},
  author={Huang, Yijiang and Garrett, Caelan R. and Ting, Ian and Parascho, Stefana and Mueller, Caitlin T.},
  journal={Construction Robotics},
  volume={5},
  pages={115--130},
  year={2021}
}

@article{wang2023temporal,
  title={A temporal coherent topology optimization approach for assembly planning of bespoke frame structures},
  author={Wang, Ziqi and Kennel-Maushart, Florian and Huang, Yijiang and Thomaszewski, Bernhard and Coros, Stelian},
  journal={ACM Trans. on Graphics},
  volume={42},
  number={4},
  year={2023}
}

@article{liang2017ras,
  title={{RAS}: a robotic assembly system for steel structure erection and assembly},
  author={Liang, Ci-Jyun and Kang, Shih-Chung and Lee, Meng-Hsueh},
  journal={International Journal of Intelligent Robotics and Applications},
  volume={1},
  pages={459--476},
  year={2017}
}
\bibliographystyle{IEEEtran}

\end{document}